# An Infinite Latent Attribute Model for Network Data

**Konstantina Palla**  KP376@CAM.AC.UK
University of Cambridge

**David A. Knowles**  DAK33@CAM.AC.UK
University of Cambridge

**Zoubin Ghahramani**  ZOUBIN@ENG.CAM.AC.UK
University of Cambridge

## Abstract

Latent variable models for network data extract a summary of the relational structure underlying an observed network. The simplest possible models subdivide nodes of the network into clusters; the probability of a link between any two nodes then depends only on their cluster assignment. Currently available models can be classified by whether clusters are disjoint or are allowed to overlap. These models can explain a "flat" clustering structure. Hierarchical Bayesian models provide a natural approach to capture more complex dependencies. We propose a model in which objects are characterised by a latent feature vector. Each feature is itself partitioned into disjoint groups (subclusters), corresponding to a second layer of hierarchy. In experimental comparisons, the model achieves significantly improved predictive performance on social and biological link prediction tasks. The results indicate that models with a single layer hierarchy over-simplify real networks.

## 1. Introduction

Network data encoding pairwise relations between objects appears in many fields. For instance, in biology, a protein network connects interacting partners, while in a social network, links among people indicate relations. We focus on the most common type of network data —sets of observations represented as an unweighted, undirected graph—in the ensuing discussion. The motivation behind the analysis of these networks is two fold. Firstly, there is a desire to understand the latent structure of the network; what are the features of the proteins that account for the observed interactions and what is the mechanism behind the links or non-links among groups of people. Second, the prediction of "missing" links in the network arises as an important challenge; how likely is it that a pair of proteins interact or that two social network members are friends. A prominent theme in machine learning is the use of latent variable methods, which approach this problem by extracting a simplified summary of the graph and predicting the presence or absence of links based on this latent representation. Latent class and latent feature models are the two most common categories found in the literature.

Latent class models assume that there are a number of clusters (classes) and that each object belongs to a single cluster. Under these models, the link probability between two objects depends only on their cluster assignments. Early work in this category includes the stochastic block model (SB) proposed in Nowicki and Snijders (2001). Instead of assuming a fixed number of clusters, the Infinite Relational Model (IRM) and the Infinite Hidden Relational Model (Kemp and Tenenbaum, 2006; Xu et al., 2006) use the Chinese restaurant process (Pitman, 2002) to allow a potentially infinite number of clusters. The Mixed Membership Stochastic Block Model (Airoldi et al., 2009) (MMSB) increases the expressiveness of the latent class models by allowing mixed membership, associating each object with a distribution over clusters.

Latent feature models increase the flexibility of the generative process by letting each object possess a vector of features and determine the link probabilities based on interactions among the features. In Hoff et al. (2001) the link probability between two objects is determined by the similarity of their real-valued fea-ture vectors. Miller et al. (2009) uses a vector of binary features which can be interpreted as allowing objects to belong to multiple clusters at the same time. Their model, the Latent Feature Infinite Relational Model (LFRM), assumes that the number of clusters is not known a priori and uses the Indian Buffet Process (IBP) (Griffiths and Ghahramani, 2005) to determine the number of latent clusters.

The limitation of a single cluster membership makes the latent class models less flexible than the latent feature models. As an intuitive example, consider a network of individuals at a collegiate University, in which a link denotes friendship or acquaintance. Here there will be multiple *types* of cluster, for instance colleges, departments and sports teams. A person might be a member of more than one cluster and his cluster-memberships determine his interaction with others. To capture this structure a single membership model, such as the IRM, must introduce a cluster for each possible combination of the types of cluster, which in our example would be to introduce clusters such as 'Gryffindor college, Department of Mathematics, Football'. This results in an exponential explosion of clusters, making learning, inference and generalisation difficult. Latent feature models, e.g. the LFRM, can instead use the feature vector representation to implicitly account for the possible combination of clusters. Though powerful, these models only account for a flat clustering of the objects. In the context of the University social network, the 'college' feature might be divided into many different subclusters, such as 'Slytherin college', 'Gryffindor college' etc. The same for 'sport', with subclusters like 'basketball', 'tennis', etc. The LFRM must represent each cluster with a new feature, which will result in feature vectors of greater size with a cost in interpretability. Allowing an explicit representation of the partitioning of each general class into subclasses would provide a more structured representation of the data.

Towards this end, we develop a new nonparametric latent feature model. We use a binary feature vector to indicate the features that an object has. If an object has a particular feature, then the object belongs to a particular subcluster of this feature. Equivalently, we can think of objects having several attributes (features) which have discrete values (the subcluster assignments). Following our university example, a person might have the 'college' attribute and belong to the 'Gryffindor college' subcluster, but cannot simultaneously be a member of another college. We denote our model by ILA for Infinite Latent Attribute model. We use a nonparametric Bayesian approach to simultaneously infer the number of features and number of sub-clusters inside each feature, while at the same time inferring what features are active for each object, which subcluster it belongs to and how subcluster membership influences the observed interactions.

The paper is arranged as follows. In Section 2 we describe the generative process for our nonparametric model. Section 3 explains the relationship of our model to several recently proposed models. In Section 4 we derive the algorithm for performing approximate posterior inference, parameter estimation and link prediction. Section 5 gives some observations about the computational cost of our proposed model relative to others. Finally, in Section 6 we study our model's performance on one synthetic and two real datasets.

## 2. Model Description

Let $\mathbf{R}$ be the $N \times N$ binary matrix that contains the links among the objects. In ILA, each object $i = 1, \ldots, N$, is represented by a binary vector of latent feature values, $\mathbf{z}_i$. If there are $M$ features, then $\mathbf{Z}$ is a $N \times M$ binary matrix indicating which features each object has active, with $z_{im} \equiv Z(i, m) = 1$ if the $i^{th}$ object has feature $m$ and $z_{im} = 0$ otherwise. Let $\mathbf{C}$ be a set of vectors, that is $\mathbf{C} = \{\mathbf{c}^{(1)}, \ldots, \mathbf{c}^{(M)}\}$, that describe the subcluster assignments within each feature, such that $\mathbf{c}^{(m)}$ is a vector of length $N$ where $c_i^{(m)}$ denotes the subcluster the $i^{th}$ object belongs to in the $m^{th}$ feature ($c_i^{(m)}$ is set to 0 if object $i$ does not have feature $m$). The number of subclusters present in the $m^{th}$ feature, which is also not known a priori, is denoted as $K^{(m)}$, so that $c_i^{(m)} \in \{0, 1, ..., K^{(m)}\}$. Finally, let $\mathbf{W}$ be a set of $M$ real-valued weight matrices of size $K^{(m)} \times K^{(m)}$ each, where $w_{kk'}^{(m)} \equiv W^{(m)}(k, k')$ is the weight that affects the probability of there being a link from object $i$ to object $j$, given that object $i$ belongs to subcluster $k$ and object $j$ belongs to subcluster $k'$ of the $m^{th}$ feature.

Given the feature matrix $\mathbf{Z}$, the set of the subcluster assignments $\mathbf{C}$, and the set of the weight matrices $\mathbf{W}$, the probability that there is a link from object $i$ to object $j$ is given by

$$\Pr(r_{ij} = 1 | \mathbf{z}_i, \mathbf{z}_j, \mathbf{C}, \mathbf{W}) = \sigma\Big( \sum_m z_{im} z_{jm} w_{c_i^m c_j^m}^{(m)} + s \Big), \quad (1)$$

where the sum ranges over all $M$ features, $s$ is a bias term, and $\sigma(x) = (1 + e^{-x})^{-1}$ is the sigmoid (logistic) function that maps the input arguments from $(-\infty, +\infty)$ to $(0, 1)$, ensuring that the result is a valid probability. Under this model, only features that are on for both objects influence the probability of a link between them. For these common features, the ap-propriate weight values are summed up, depending on the subcluster assignments of $i$ and $j$. The weight values are continuous variables which can be positive or negative allowing pairs of subclusters to encourage or discourage links between them correspondingly. We assume that given the $\mathbf{Z}$, $\mathbf{C}$ and $\mathbf{W}$, the probability of each link is independent and the likelihood is therefore as follows

$$\Pr(\mathbf{R}|\mathbf{Z},\mathbf{C},\mathbf{W}) = \prod_{i,j} \Pr(r_{ij}|\mathbf{z}_i, \mathbf{z}_j, \mathbf{C}, \mathbf{W}). \quad (2)$$

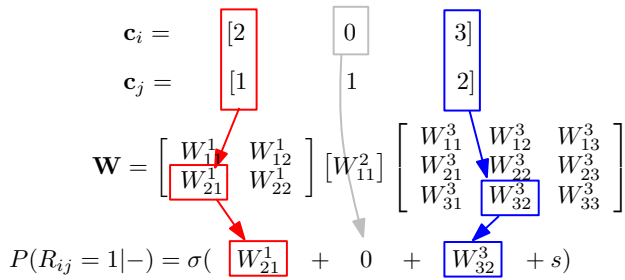

Figure 1. Diagram of the ILA model. $\mathbf{c}_i$ and $\mathbf{c}_j$ are the sub-cluster assignments for objects $i$ and $j$ respectively, shown here with $M = 3$ features. $c_i^{(2)}$ being zero corresponds to the absence of feature 2 for object $i$, so this feature contributes no weight. For the two features which are active for both $i$ and $j$, namely features 1 and 3, the subcluster assignments dictate which element of the feature's weight matrix should be chosen for each feature. Finally the weights are summed and passed through a sigmoid function to give the probability of a link between $i$ and $j$.

In order to allow flexible inference of the latent structure from data, we set the number of possible features $M$ and the number of subclusters in each feature $K^{(m)}$ to infinity by using an IBP prior on $\mathbf{Z}$ and CRP priors on the $\mathbf{c}$'s. The hierarchical generative model is then:

$$\mathbf{Z}|\alpha \sim \text{IBP}(\alpha)$$
$$\mathbf{c}^{(m)}|\gamma \sim \text{CRP}(\gamma)$$
$$w_{kk'}^{(m)}|\sigma_w \sim N(0, \sigma_w^2)$$
$$r_{ij}|\mathbf{Z},\mathbf{C},\mathbf{W} \sim \text{Bernoulli}\left(\sigma\left(\sum_m z_{im} z_{jm} w_{c_i^m c_j^m}^{(m)} + s\right)\right).$$

The ILA model is illustrated in Figure 1.

The IBP parameter, $\alpha$, affects the number of represented features, whereas the CRP parameter, $\gamma$, controls the number of subclusters inside each feature. To improve the flexibility of our model, we put Gamma priors on $\alpha$ and $\gamma$, and a Gaussian prior on the bias term $s$ as follows

$$\alpha \sim \mathcal{G}(1,1), \qquad \gamma \sim \mathcal{G}(1,1), \qquad s \sim \mathcal{N}(\mu_s, \sigma_s^2),$$

where $\mu_s$ and $\sigma_s$ are the mean and standard deviation hyperparameters for the bias (we use $\mu_s = -1, \sigma_s = 4$ unless otherwise stated).

## 3. Related work

Here we examine three models that are closely related to ILA. The IRM model of Kemp and Tenenbaum (2006) and the LFRM of Miller et al. (2009) both use nonparametric Bayesian approaches to account for potentially infinite number of clusters in the data. In the IRM, the link probability between two objects depends only on the clusters they are assigned to:

$$\Pr(r_{ij} = 1|c_i, c_j, \eta) = \eta_{c_i c_j}, \quad (3)$$

where the link probabilities for each pair of clusters, $\{\eta_{kk'} : k, k' = 1, ..., K\}$ are given independent Beta priors, and the cluster assignments, $c$ are given a CRP prior. The ILA and LFRM on the other hand, put a logistic-normal prior on the between feature and subcluster link probabilities. More specifically, the LFRM defines the link probability as

$$\Pr(r_{ij} = 1|\mathbf{Z},\mathbf{W}) = \sigma\left(\sum_{kl} z_{ik} z_{jl} w_{kl} + s\right), \quad (4)$$

where $\mathbf{W}$ is a $K \times K$ real valued weight matrix (with $K$ being the number of features), given an element-wise Gaussian prior, and $\mathbf{Z}$ is an $N \times K$ matrix of binary feature vectors drawn from an IBP. Comparing Equations 4 and 1 for the ILA, we see how the two models differ. The LFRM defines a weight value for each possible pair of features, while ILA defines a weight *matrix* for each feature, whose elements correspond to every pair of subclusters in that feature. The link probability in LFRM depends on all the possible pairs of features that are on for both objects, while in the ILA model, the link probability is contributed to only by features that are *simultaneously* on for both objects. While subclusters within a feature can interact in ILA, subclusters from different features do not interact.

Unlike the IRM, the ILA model does not partition the objects into a set of non-overlapping clusters; although it specifies non-overlapping subclusters for each feature, it also allows each object to have multiple features, thus accounting for multiple membership. ILA more expressive than LFRM because it associates each feature with a set of subclusters.

Interestingly, both the IRM and LFRM can be thought of as special cases of our model. If only one column of $\mathbf{Z}$ is switched on in ILA (i.e. there is only one feature which is on for every object) then this is equivalent to the IRM. In this case the ILA likelihood becomes

$$\Pr(r_{ij} = 1|\mathbf{Z} = \mathbf{1}, \mathbf{C}, \mathbf{W}) = \sigma\left(w^{(1)}_{c_i^1 c_j^1} + s\right). \quad (5)$$

Contrasting this to Equation 3 the ILA has a logistic-normal prior on the between subcluster link probabilities rather than a Beta prior, but this is a relatively minor difference.

If the LFRM is constrained to have a weight matrix $\mathbf{W}$ with only diagonal non-zero elements, then its link probability becomes

$$\Pr(r_{ij} = 1|\mathbf{Z}, \mathbf{W}) = \sigma\left(\sum_k z_{ik} z_{jk} w_{kk} + s\right).$$

This is then equivalent to ILA in the case when there is only one subcluster in each feature, since the ILA link probability is then

$$\Pr(r_{ij} = 1|\mathbf{z}_i, \mathbf{z}_j, \mathbf{C} = \mathbf{1}, \mathbf{W}) = \sigma\left(\sum_m z_{im} z_{jm} w^{(m)}_{11} + s\right).$$

The ILA model can also be seen as a extension of the Multiplicative Attribute Graph (MAG) model proposed in Kim and Leskovec (2011), where the link probability is

$$\Pr(r_{ij} = 1|\mathbf{C}, \eta) = \prod_m \eta^{(m)}_{c_i^m c_j^m},$$

where $\eta$ is a set of $M$ two by two matrices of probabilities with elementwise independent Beta priors, and the $\mathbf{c}$'s are equivalent to our subcluster assignment variables but constrained to takes values in $\{1,2\}$. We extend this model in three ways: 1) we learn the number of subclusters in each feature, rather than fixing it to two, 2) we learn the number of features $M$, and 3) we incorporate additional sparsity, in that an object need not have a particular feature active at all. We parameterise our model in terms of real valued weights which contribute to the log odds of a link being on, rather than with probabilities that are multiplied together, but this entails no loss of flexibility. In fact this may be advantageous to ILA since the MAG suffers from each new feature decreasing all link probabilities.

There are several models that have been proposed for discovering hierarchical structure in relational data (Girvan and Newman, 2002; Roy et al., 2007). In these models, each object is still a member of one out of many non-overlapping clusters. Our model is distinct in allowing each object to be a member of many subclusters as long as these subclusters are in different features.

## 4. Inference

In the following, we present a method for inferring the latent variables of the model: the infinite binary feature matrix $\mathbf{Z}$, the subcluster assignments, $\mathbf{c}^{(m)}$ for each feature $m$, and the weight matrices, $\mathbf{W}^{(m)}$. Simultaneously we recover the number of features and the number of subclusters inside each feature. As with many other Bayesian models, exact inference is intractable so we employ Markov Chain Monte Carlo (MCMC), and follow an iterative procedure that achieves posterior inference over the latent variables. The sampler iterates as follows:

**Sampling the feature matrix, $\mathbf{Z}$.** We Gibbs sample each element of $\mathbf{Z}$ in succession. For each object $i$, the sampler makes the following decisions: which of the current $M$ available features should be turned on/off, and how many new features should be turned on. However, when turning on a feature the sampler must also sample a new subcluster assignment and, in case of adding a new subcluster, the related new weights.

We use exchangeability of the rows of $\mathbf{Z}$ and assume that the $i^{th}$ object is the last to be added to $\mathbf{Z}$ after $N-1$ rows have already been added. For all the $M$ features currently present in $\mathbf{Z}$, the conditional posterior probability of an entry $z_{im}$, $m = 1, \ldots, M$ follows a Bernoulli distribution:

$$\Pr(z_{im} = 1|\mathbf{Z}_{-im}, \mathbf{C}_{-im}, \mathbf{W}, \mathbf{R}) \propto$$
$$\frac{n_{-im}}{N}\Pr(\mathbf{R}|z_{im} = 1, \mathbf{Z}_{-im}, \mathbf{C}_{-im}, \mathbf{W}), \quad (6)$$

where $\mathbf{Z}_{-im}$ is the $\mathbf{Z}$ matrix excluding the $Z(i,m)$ element, $n_{-im}$ is the number of times feature $m$ is present in $\mathbf{Z}_{-im}$ and $\mathbf{C}_{-im}$ excludes the subcluster assignment $c_i^{(m)}$. To compute the probability in Equation 6, we need to sum over $c_i^{(m)}$, the space of the possible subclusters that the $i^{th}$ object may be assigned to if $z_{im}$ is to be turned on. This also includes integration over a possible new subcluster. However, the prior over the parameters $\mathbf{W}^{(m)}$ related to a new subcluster is not conjugate because of the logistic link function, and thus the likelihood term cannot be computed exactly. To overcome this problem, we use the auxiliary variable approach proposed in Neal (2000) (Algorithm 8), both to facilitate the integration required in Equation 6, and to decide which subcluster to assign the $i^{th}$ object to in the $m^{th}$ feature if $z_{im}$ is turned on.

We must also sample the number of new features unique to the $i^{th}$ row, $M^{(i)}_{\text{new}}$. Instead of considering these features separately, we calculate the conditional posterior over $M^{(i)}_{\text{new}}$, using the fact that under the IBP the prior distribution over $M_\text{new}$ for the last row is Poisson($\alpha/N$). Combining the Poisson prior with the likelihood, we obtain the conditional posterior over $M_\text{new}^{(i)}$. However, to obtain the required likelihood term we need values for $\mathbf{C}^{(m)}$ and $\mathbf{W}^{(m)}$ for the proposed new features. Clearly $c_i^{(m)} = 1$ for any new features, since a feature active for only one object can only have one subcluster. Integrating over the weights is not straightforward because the prior over $\mathbf{W}^{(m)}$ is not conjugate to the logistic likelihood. We therefore employ a Metropolis Hastings step, proposing values for $w_{11}^{(m)}$ from the prior so that the acceptance ratio becomes simply the likelihood ratio for including the new features and associated $\mathbf{C}^{(m)}$ and $\mathbf{W}^{(m)}$ values in the model versus not including them.

**Sampling the subcluster assignments, C.** We may choose to resample each $\mathbf{C}^{(m)}$ in succession as a second step, again using Algorithm 8 of Neal (2000). In practice we found this unnecessary since $\mathbf{C}$ is sampled in the process of sampling $\mathbf{Z}$.

**Sampling the weights, W.** Given $\mathbf{Z}$ and $\mathbf{C}$, the sampler successively resamples each of the weights $\{w_{kk'}^{(m)} : k, k' = 1, \ldots, K^{(m)}, m = 1, \ldots, M\}$. Since we do not have conjugacy (due to the logistic link function), we cannot sample directly from the posterior over $w_{kk'}^{(m)}$. To overcome this problem we used both Metropolis Hastings and slice sampling (Neal, 2003) but found the later resulted in faster mixing.

**Hyperparameters.** We use slice sampling for both the IBP hyperparameter, $\alpha$, the CRP concentration parameter, $\gamma$ and the bias, $s$.

**IRM implementation.** Our implementation of the IRM model of Kemp and Tenenbaum (2006) uses standard single site Gibbs sampling along with the restricted Gibbs sampling split merge method of Jain and Neal (2000). In the IRM we are able to integrate out the parameters $\eta$ analytically due to conjugacy, so we need only sample the cluster assignments and the CRP concentration parameter, for which we use slice sampling.

**LFRM implementation.** For the LFRM of Miller et al. (2009), we Gibbs sample the IBP matrix $\mathbf{Z}$ and slice sample each element of the weight matrix $\mathbf{W}$ sequentially, followed by the IBP concentration parameter.

### 4.1. Sequential initialisation

The Gibbs updates described above are the simplest moves we could make in a MCMC inference procedure for the ILA model. However, these updates are quite incremental, since only a single variable is updated at a time. Due to the extremely large number of possible configuration states, $\prod_{m=1}^{M}(K^{(m)}+1)^N$, the sampler can suffer from local modes and have somewhat slow mixing. Non-incremental moves, like splitting and merging features in the $\mathbf{Z}$ matrix or subcluster assignments in $\mathbf{C}$ can produce major changes in the configuration state in a single iteration and can help the sampler explore more efficiently. Split-merge sampling in the IBP has been previously described in Meeds et al. (2006). However, we found that a sequential initialization of the sampler improved the performance, guiding the sampler closer to neighborhoods of higher probability.

To sequentially initialise all parameters the objects are first randomly permuted and then added to the model as follows. Initially two objects are added to the model with no features active. Then a few (typically three) iterations of the MCMC sampler are run. Then the next object is added, with no features turned on, and another three iterations of the sampler are run. This procedure is iterated until all objects have been added. The sampler will naturally grow the number of features and subclusters within each feature as more data is added. The advantage of this method is that the initialisation is appropriate for the model, the sampler is very fast initially due to the small number of objects, and the search space is small initially so it is easier for the Markov chain to find a relatively high probability region of parameter space. We also used sequential initialisation for our implementation of LFRM, but not for IRM where we find split-merge sampling is able to better overcome local optima.

### 4.2. Prediction

A principled way to evaluated a generative model is by its ability to predict missing data values given some observations. In our model, we collect $T$ samples $\{\{\mathbf{Z}_{(1)}, \mathbf{C}_{(1)}, \mathbf{W}_{(1)}\}, \ldots, \{\mathbf{Z}_{(T)}, \mathbf{C}_{(T)}, \mathbf{W}_{(T)}\}\}$ and estimate the predictive distribution of a missing link as the average of the predictive distributions for each of the collected samples. Assuming that we want to predict the missing link $r_{ij}$ between objects $i$ and $j$, the approximate predictive distribution will be as follows

$$\Pr(r_{ij} = 1 | \mathbf{R}_\text{train}) \approx \frac{1}{T} \sum_{t=1}^{T} \Pr(r_{ij} = 1 | \mathbf{Z}_{(t)}, \mathbf{C}_{(t)}, \mathbf{W}_{(t)}).$$

An Infinite Latent Attribute Model for Network Data

## 5. Computational complexity

In general, the computational cost of latent feature models scales quadratically in the number of objects. In the LFRM, computing the likelihood has a complexity of $\mathcal{O}(M^2N^2)$, where $M$ and $N$ is the number of represented features[1] and the number of objects correspondingly. For ILA, the link probability between two objects given by Equation 1, results in computational cost $\mathcal{O}(MN^2)$ when calculating the likelihood across all pairs. The computational cost of the IRM scales linearly in the number of links in the network, $L = \sum_{ij} r_{ij}$, because the likelihood, with the link probabilities $\eta$ integrated out, can be written as

$$\Pr(\mathbf{R}|\mathbf{c}) = \prod_{a,b} \frac{\text{Beta}(n(a,b)+\beta, \bar{n}(a,b)+\beta)}{\text{Beta}(\beta,\beta)},$$

where $n(a,b)$ is the number of pairs of objects $(i,j)$ where $i \in a$ and $j \in b$ and $R(i,j) = 1$, $\bar{n}(a,b)$ is the number of such pairs where $R(i,j) = 0$, and $\text{Beta}(\cdot,\cdot)$ is the Beta function. The computational cost of computing the likelihood in the IRM is therefore $\mathcal{O}(K^2L)$.

In Morup et al. (2011), it is observed that if a noisy-or likelihood model is used in the LFRM rather than the logistic Gaussian model, then the likelihood can be calculated in $\mathcal{O}(K^2L)$ as for the IRM. This allows excellent scalability on typical sparse real world networks where the number of links is much smaller than the number of non-links. This scalable variant is applicable to our model, but comes with the significant restriction of only being able to have positive weights between clusters (homophily). As a result we leave this development to future work.

## 6. Results

We present results on a toy synthetic data set and on two real world datasets: the NIPS coauthorship network and a novel gene interaction network.

### 6.1. Synthetic data

We first explored the ability of our model to recover the underlying structure of a network using synthetic data. We considered one simple synthetic dataset (Figure 2a) hand-constructed to have an unambiguous most parsimonious solution under each model. Under ILA this is the feature matrix shown in Figure 2(b) with two features. The first feature has three homophilic subclusters (i.e. individuals tend to have links if they are in the same cluster), whereas the second feature has two heterophilic subclusters (i.e. individuals tend to link if they are in different clusters). We ran ILA for 200 MCMC iterations following sequential initialisation. The sample with the lowest energy (highest log probability under the posterior) corresponds exactly to the expected "true" structure, as shown in Figure 2b. The MAP sample found using LFRM is shown in Figure 2c. Again this is a passable explanation of the data but it is considerably more convoluted than the simple, interpretable but rich solution found using ILA. Note that running 2000 iterations (following sequential initialisation) of LFRM no better solution was found. In contrast the IRM finds the flat clustering of six clusters shown in Figure 2d, which is an acceptable solution but does not capture the rich structure that ILA is able to.

[1]$M$ is potentially unbounded, but in practice the model will use some finite number of features to model any finite dataset.

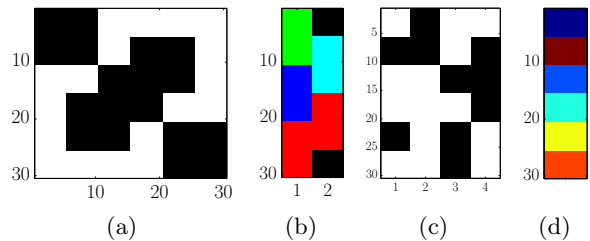

Figure 2. Synthetic data example. (a) Observed synthetic $30 \times 30$ link matrix. White corresponds to zero, black to one. (b) ILA solution. Rows correspond to individuals, columns correspond to features. Different colors correspond to different subclusters. Black denotes that this feature is inactive. (c) LFRM solution. White corresponds to zero, black to one (active feature). (d) IRM solution. Different colours denote the different cluster assignments.

### 6.2. NIPS coauthorship network

We compare the performance of the IRM, LFRM and ILA on the NIPS coauthorship dataset (Globerson et al., 2007), where a link corresponds to two individuals being coauthors of a paper at one of the first 17 NIPS conferences (see Figure 3(a)). Following Miller et al. (2009) we use only the 234 most connected authors. We run 10 repeats, each time holding out a different 20% of the data (links and non-links) and using a different random initialisation. We run two versions of ILA: the first with a fixed number of features $M = 6$, and the second which learns $M$, denoted $M = \infty$. Note that even with $M = 6$ ILA is still extremely flexible since it can learn the number of subclusters in each feature. We run 500 iterations for ILA and 1000 iterations for IRM and LFRM, and calculate evaluation metrics averaged over the last 300 samples. The results are shown in Table 1. We confirm the finding in Miller et al. (2009) that LFRM outperforms the IRM on this dataset. However, across all three evaluation metrics both ILA versions significantly outperform LFRM (for example, the t-test between the test log likelihoods for LFRM and ILA ($M = 6$) shows the means to be significantly different with a *p*-value of $10^{-7}$). The fully infinite version of ILA performs slightly, but still statistically significantly, better than when we constrain $M = 6$ for training error and test error. For test log likelihood ILA ($M = \infty$) still appears to perform slightly better than ILA ($M = 6$) but the difference is not statistically significant based on a t-test. Under the ILA posterior $M$ is concentrated around 7 or 8 features, with typically 2 to 4 subclusters per feature.

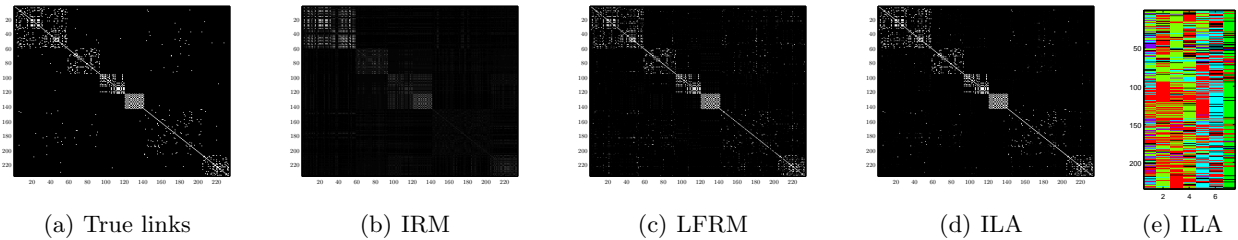

(a) True links     (b) IRM     (c) LFRM     (d) ILA     (e) ILA

*Figure 3.* Predictions for the three models on the NIPS 1-17 coauthorship dataset. In (a), white denotes that two people wrote a paper together, while in (b)-(d), the lighter the entry, the more confident the model is that the corresponding authors would collaborate. In (e), we present the subclusters recovered by ILA in the 7 corresponding features. Different colors denote the different subcluster assignments.

*Table 1.* NIPS coauthorship network results. All values are averages over the test pairs. The best results are highlighted in bold where statistically significant.

|  | IRM | LFRM | ILA ($M = 6$) | ILA ($M = \infty$) |
|---|---|---|---|---|
| Train error (0-1 loss) | $0.0427 \pm 0.0009$ | $0.0197 \pm 0.0052$ | $0.0086 \pm 0.0005$ | $\mathbf{0.0058 \pm 0.0005}$ |
| Test error (0-1 loss) | $0.0440 \pm 0.0014$ | $0.0228 \pm 0.0041$ | $0.0141 \pm 0.0012$ | $\mathbf{0.0106 \pm 0.0007}$ |
| Test log likelihood | $-0.0859 \pm 0.0043$ | $-0.0547 \pm 0.0079$ | $\mathbf{-0.0322 \pm 0.0058}$ | $-0.0318 \pm 0.0094$ |
| AUC | $0.9565 \pm 0.0037$ | $0.9631 \pm 0.0150$ | $\mathbf{0.9908 \pm 0.0048}$ | $\mathbf{0.9910 \pm 0.0056}$ |

In Figure 3 the link predictions for each of the three models are presented. Figures 3(b)-(d) visualize the belief of each model that there should be a link between each pair of authors. The link matrices were constructed after running the three models on the NIPS 1-17 dataset for 500 iterations, using the same random seed and averaging over the last 160 samples. To facilitate interpretability, we ordered the authors by the clusters found by the IRM. It can be clearly seen that both the LFRM and ILA models outperform the IRM model by appearing more confident and reproducing the corresponding network more faithfully.

Considering Figures 3(c)-(d), LFRM and ILA appear comparable, with ILA being slightly more confident. Quantitatively however, ILA gives a test log likelihood of $-0.0295$ as opposed to $-0.0386$ for the LFRM model. We also report the AUC metric, the area under the ROC (Receiver Operating Characteristic) curve, for the held-out data.

### 6.3. Gene interaction network

Finally we present results on a subset of the interaction data presented in Jonikas et al. (2009)[2]. This is an example of a new class of high throughput gene interaction assays, in this case using the yeast *S. cerevisiae*. A range of "deletion" strains are created, each of which has a single gene deleted. Some phenotypic response is measured during the growth of each strain, in this case unfolded protein response (UPR), a measure of how badly the cell is doing at correctly folding its membrane proteins. "Double mutants" with two distinct genes deleted are then screened. Based on the single deletion strains, the expected UPR response for these double mutants can be predicted (see Jonikas et al. (2009) for details) assuming no interaction between the two deleted genes. If the observed UPR response is significantly different from this predicted value then the genes must interact in some way, so we consider this as an edge in the network. We use the 156 genes with the least missing data. We run 10 repeats with a different 10% of the observed data held-out each time, and perform 500 MCMC iterations for ILA and 1000 for the IRM and LFRM. Again we find the ILA model significantly outperforms LFRM, which in turn outperforms the simple IRM (see Table 2). In this case the infinite version $M = \infty$ has considerably better predictive performance than with $M = 6$, suggesting there is considerably more structure in this data so that allowing more features is beneficial. In fact ILA typically finds around $M = 30$ features with 3 to 5 subclusters per feature. We find significantly more features are associated with particular properties of the genes as defined by Gene Ontology classes[3] than would be expected by chance ($p < 10^{-3}$ calculated by permutation testing), for example the three subclusters of one particular feature have very different proportions of ligand binding genes ($10/41, 21/27$ and $2/20$ respectively).

---

[2]See http://weissmanlab.ucsf.edu/upremap/

*Table 2.* Gene interaction network results. All values are averages over the test pairs. The best results are highlighted in bold where statistically significant.

|  | IRM | LFRM | ILA ($M = 6$) | ILA ($M = \infty$) |
| --- | --- | --- | --- | --- |
| Train error (0-1 loss) | $0.3562 \pm 0.0008$ | $0.2603 \pm 0.0098$ | $0.2044 \pm 0.0066$ | $\mathbf{0.0248 \pm 0.0010}$ |
| Test error (0-1 loss) | $0.3608 \pm 0.0031$ | $0.2661 \pm 0.0086$ | $0.2284 \pm 0.0077$ | $\mathbf{0.0735 \pm 0.0047}$ |
| Test log likelihood | $-0.4669 \pm 0.0097$ | $-0.4223 \pm 0.0147$ | $-0.3596 \pm 0.0156$ | $\mathbf{-0.2654 \pm 0.0447}$ |
| AUC | $0.8654 \pm 0.0057$ | $0.8471 \pm 0.0132$ | $0.9401 \pm 0.0046$ | $\mathbf{0.9924 \pm 0.0037}$ |

## 7. Conclusion

Our experimental results on two very different datasets suggest that the network models proposed to date fail to capture the complex nature of real world networks. We have introduced a hierarchical nonparametric Bayesian model, ILA, which is able to naturally represent this complexity, with corresponding gains in empirical performance. In principle ILA could be made even more flexible by allowing multiple membership of subclusters within a feature, corresponding to a hierarchical IBP. We leave investigating whether this is beneficial to future work.

---

[3] http://www.geneontology.org/